\documentclass{arxivtmpl}

\def\eqref#1{equation~\ref{#1}}

\def\1{\bm{1}}

\DeclareMathAlphabet{\mathsfit}{\encodingdefault}{\sfdefault}{m}{sl}
\SetMathAlphabet{\mathsfit}{bold}{\encodingdefault}{\sfdefault}{bx}{n}

\definecolor{darkred}{RGB}{140, 21, 21}
\definecolor{lightgray}{gray}{0.7}
\definecolor{orange}{HTML}{F58025}
\definecolor{deepred}{rgb}{0.631,0.102,0.102}
\definecolor{amethyst}{rgb}{0.6, 0.4, 0.8}
\definecolor{darkgreen}{rgb}{0.3,0.7,0.3}
\definecolor{salmon}{RGB}{241, 150, 141}
\definecolor{mildyellow}{HTML}{FFF2CC}

\hypersetup{
    colorlinks=false,
    pdfborderstyle={/S/U/W 0.5},
    pdfborder={0 0 0.5},
    linkbordercolor=orange,
    citebordercolor=orange,
    filebordercolor=orange,
    urlbordercolor=orange
}

\renewcommand\Affilfont{\normalsize\normalfont}

\definecolor{mygreen}{HTML}{3cb44b}
\definecolor{skyblue}{HTML}{beffff}
\definecolor{lightgreen}{HTML}{90ee90}

\definecolor{emerald}{rgb}{0.31, 0.78, 0.37}

\newdateformat{ymd}{\the\year-\the\month-\the\day}

\definecolor{mygreen}{HTML}{3cb44b}
\colorlet{myyellow}{green!10!orange!90!}
\makeatletter

\usetikzlibrary{arrows,shapes,snakes,automata,backgrounds,fit,petri}

\newcommand{\RN}[1]{%
	\textup{\lowercase\expandafter{\it \romannumeral#1}}%
}

\newcommand{\beq}{\vspace{0mm}\begin{equation}}
\newcommand{\eeq}{\vspace{0mm}\end{equation}}
\newcommand{\beqs}{\vspace{0mm}\begin{eqnarray}}
\newcommand{\eeqs}{\vspace{0mm}\end{eqnarray}}
\newcommand{\barr}{\begin{array}}
\newcommand{\earr}{\end{array}}

\definecolor{Gray}{gray}{0.93}

\definecolor{mygreen}{HTML}{3cb44b}

\newcommand{\dat}[1]{#1}

\hypersetup{pdftitle={OpenDeepThink: Parallel Reasoning via Bradley--Terry Aggregation}}

\title{OpenDeepThink: Parallel Reasoning via Bradley--Terry Aggregation}

\author[1]{Shang Zhou}
\author[2]{Wenhao Chai}
\author[3]{Kaiyuan Liu}
\author[4]{Huanzhi Mao}
\author[4]{Qiuyang Mang}
\author[1]{Jingbo Shang$^{\dagger}$}
\affil[1]{UC San Diego}
\affil[2]{Princeton University}
\affil[3]{University of Washington}
\affil[4]{UC Berkeley}
\authornote{$^{\dagger}$ Advising author}

\makeatletter
\newcommand{\pagehead}{\@title}
\makeatother

\begin{document}

\maketitle
\thispagestyle{firstpageblank}

\begin{abstract}
Test-time compute scaling is a primary axis for improving LLM reasoning.
Existing methods primarily scale depth by extending a single reasoning trace. Scaling breadth by sampling multiple candidates in parallel is straightforward, but introduces a selection bottleneck: choosing the best candidate without a ground-truth verifier, since pointwise LLM judging is noisy and biased.
To address this, we introduce OpenDeepThink, a population-based test-time compute framework that selects via pairwise Bradley--Terry comparison.
Each generation, the LLM judges random pairs of candidates and aggregates votes via Bradley--Terry into a global ranking; top-ranked candidates are preserved and the top three quarters are mutated using the natural-language critiques produced during comparison; the bottom quarter is discarded.
OpenDeepThink raises Gemini~3.1~Pro's effective Codeforces Elo by $+405$ points in eight sequential LLM-call rounds (${\sim}27$ minutes wall-clock).
The pipeline transfers across weaker and stronger models without retuning, and on the multi-domain HLE benchmark, gains appear concentrated in objectively verifiable domains and reverse in subjective ones.
We release CF-73, a curated set of 73 expert-rated Codeforces problems with International Grandmaster annotation and $99\%$ local-evaluation agreement against the official verdict.\footnote{\url{https://github.com/ZhouShang0817/CF-73}}
\end{abstract}

\section{Introduction}
\label{sec:intro}

Reasoning models such as o1~\citep{jaech2024o1} and DeepSeek-R1~\citep{guo2025deepseekr1} have established test-time compute scaling as a primary axis for improving LLM reasoning.
The gains are sharp on mathematical reasoning, competitive programming, and agentic benchmarks.
The dominant paradigm extends the model's chain of thought, either by encouraging longer traces~\citep{jaech2024o1,guo2025deepseekr1} or by searching over reasoning steps with a learned value function~\citep{snell2024scaling,wu2024inference}.
Both approaches are inherently sequential: additional compute buys depth, not breadth, and a single early misstep derails the rest of the trace~\citep{huang2024selfcorrect}.
Best-of-$N$ sampling parallelizes naturally but shifts the bottleneck to selection.
Picking the best candidate from a pool requires either ground-truth test cases~\citep{cobbe2021training}, a trained reward model~\citep{lightman2024lets,wang2024mathshepherd}, or an LLM judge whose pointwise scores are noisy and positively biased~\citep{zheng2023judging,liu2023geval}.
Self-refinement methods~\citep{madaan2023selfrefine,shinn2023reflexion} iterate on a single trajectory and lack population-level selection pressure, tending to patch a fixed approach rather than revise it~\citep{huang2024selfcorrect}.
To our knowledge, no existing method simultaneously parallelizes across candidates and combines selection with directed mutation without requiring domain-specific verification infrastructure.

\begin{figure}[t]
  \centering
  \includegraphics[width=\linewidth]{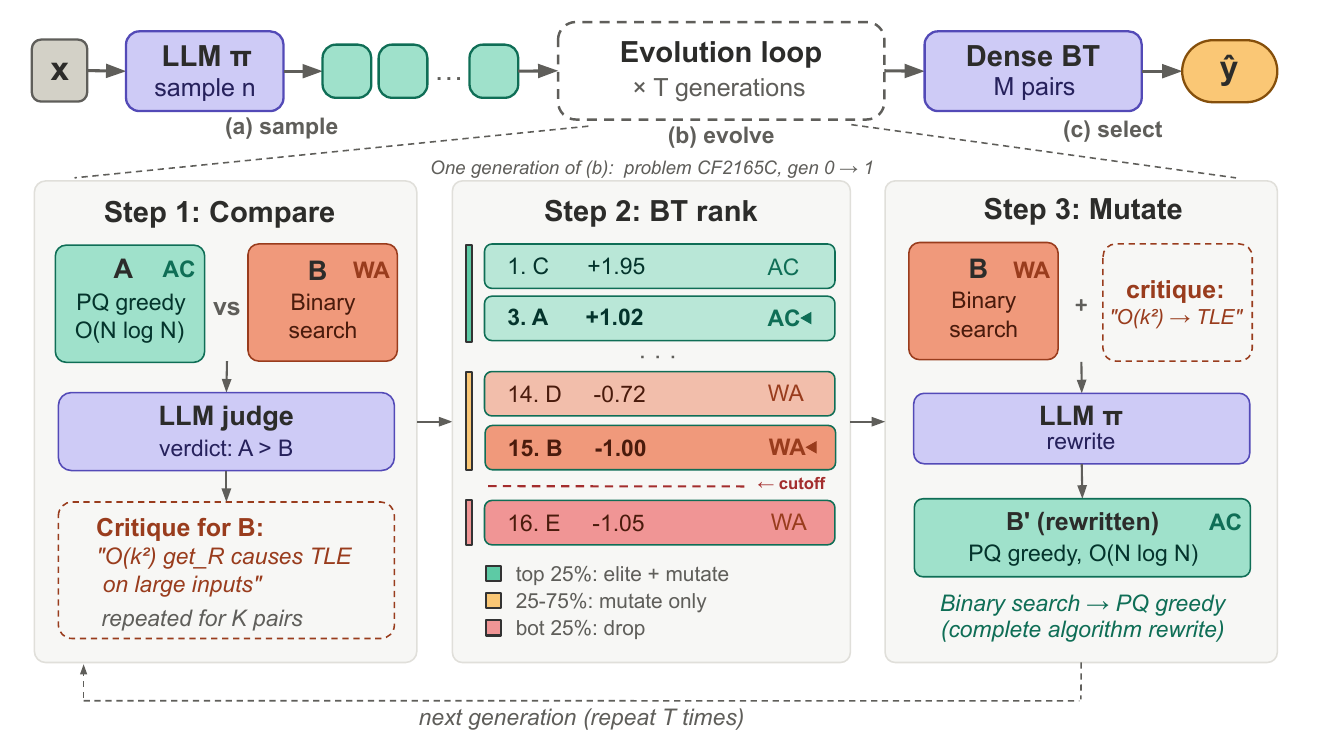}
  \caption{OpenDeepThink pipeline. (a)~Initial parallel sampling of $n$ candidate solutions. (b)~An evolution loop of $T$ generations, each performing $K$ randomized pairwise comparisons per candidate, Bradley--Terry aggregation into a global ranking, top-quartile elite preservation, feedback-driven mutation of the top $75\%$ (including elites), and discarding of the bottom quartile. (c)~A final round of $M$ denser pairwise comparisons feeds a Bradley--Terry ranking that selects the submitted solution. All three steps are embarrassingly parallel across candidates and require no external verifier. Verdict labels (AC/WA) are shown for post-hoc analysis only; the pipeline operates without access to any ground-truth signal.}
  \label{fig:pipeline}
\end{figure}

Together, these three properties point to a population-based design rather than a single trajectory.
A population enables head-to-head ranking without external verifiers, and the losers can be replaced via directed mutation rather than patched.
OpenDeepThink realizes this design: a parallel test-time compute framework that maintains a population of $n$ candidate solutions and evolves them over $T$ generations.
Each generation performs three steps: (i)~randomized pairwise comparisons judged by the same LLM that generated the candidates, (ii)~Bradley--Terry aggregation~\citep{bradley1952rank} of the comparison outcomes into a global ranking, and (iii)~feedback-driven mutation of non-discarded candidates, conditioned on the natural-language critiques from comparison.
Top-ranked candidates are preserved as elites and, together with the middle ranks, regenerated under targeted feedback that permits abandoning the current approach entirely; the bottom quarter is discarded.
A final, denser round of pairwise comparisons feeds a Bradley--Terry ranking that selects the submitted solution.
The entire pipeline requires approximately 285 API calls per problem, with a sequential depth of only eight LLM calls.
All calls within each round execute in parallel.

The core mechanism is pairwise Bradley--Terry comparison.
Pairwise framing matters because pointwise LLM judging is positively biased~\citep{zheng2023judging,liusie2024pairwise}.
On a controlled 500-pair diagnostic, pairwise reaches $86\%$ accuracy versus $59\%$ for pointwise.

We evaluate OpenDeepThink on two competition-level programming benchmarks totaling 192 problems and on 82 questions from the multi-domain HLE benchmark~\citep{phan2025hle}, using Gemini~3.1~Pro as both generator and judge. Our contributions are as follows. (i)~We introduce OpenDeepThink, a population-based test-time compute framework that combines pairwise Bradley--Terry selection with feedback-driven mutation, requiring no external verifier or domain-specific infrastructure. (ii)~On competitive programming, the framework raises Gemini~3.1~Pro's effective Codeforces Elo by $+405$ points, comparable to the $+411$ that Gemini~3~Deep Think achieves over Gemini~3.1~Pro on LiveCodeBench Pro~\citep{zheng2025livecodebenchpro}; the same hyperparameters transfer to Gemini~3~Flash and Gemini~2.5~Pro without retuning. (iii)~On HLE, gains appear concentrated in objectively verifiable domains and reverse in subjective ones, suggesting that the framework's effectiveness tracks the reliability of pairwise LLM judgment. (iv)~We release CF-73, a curated set of 73 expert-annotated Codeforces problems with $99\%$ local-evaluation agreement against the official verdict.

\section{Related Work}
\label{sec:related}

\paragraph{Test-time compute scaling and verifier-based selection.}
The dominant approach to improving LLM reasoning at inference scales compute along a single sequential trace, either by training models to produce longer chains of thought~\citep{jaech2024o1,guo2025deepseekr1} or by searching over reasoning steps with a learned value function~\citep{snell2024scaling}. Self-consistency~\citep{wang2023selfconsistency} parallelizes naturally by sampling multiple traces and selecting the majority answer, but is limited to tasks with extractable, votable final answers. When the output is a full program or an open-ended argument, selection requires a stronger signal: best-of-$N$ methods rely on trained outcome or process reward models~\citep{cobbe2021training,lightman2024lets,wang2024mathshepherd}, and tree-search methods require a value function over partial reasoning states~\citep{yao2023tree,hao2023rap}. \citet{brown2024monkeys} show that coverage scales log-linearly with sample count but that existing selectors plateau without a ground-truth verifier, articulating the exact bottleneck OpenDeepThink targets. Our framework removes the verifier requirement entirely: it replaces trained reward models and majority voting with a Bradley--Terry ranking over pairwise LLM judgments, an aggregator that handles open-ended outputs and internalizes opponent strength without any labeled data.

\paragraph{Self-refinement.}
Self-Refine~\citep{madaan2023selfrefine} and Reflexion~\citep{shinn2023reflexion} iterate on a single trajectory through feedback--rewrite cycles, improving outputs without additional training. However, \citet{huang2024selfcorrect} demonstrate that intrinsic self-correction, without external feedback or oracle stopping, often degrades performance, because a single trace cannot reliably identify its own errors. OpenDeepThink responds directly to this finding: errors are surfaced between candidates via pairwise comparison, an easier discriminative task than absolute self-evaluation, and only the aggregated signal drives mutation. \citet{dang2026cognitivewell} independently identify this failure mode as the ``Cognitive Well'': in solver--grader pipelines, iterative refinement can converge to a confident but incorrect solution that the pipeline's own grader cannot reject, motivating their use of conjecture extraction and independent verification rather than in-place refinement.

\paragraph{Evolutionary methods and LLM-as-judge.}
OpenDeepThink grafts two lines of work onto a shared scaffold. On the evolutionary side, FunSearch~\citep{romeraparedes2024funsearch} and AlphaEvolve~\citep{novikov2025alphaevolve} use LLMs as mutation operators over populations of programs but depend on programmatic, ground-truth evaluators for fitness; EvoPrompt~\citep{guo2024evoprompt} evolves prompts rather than solutions and requires a labeled development set. OpenDeepThink inherits the population--mutation--selection loop but replaces the evaluator with a training-free fitness signal: pairwise LLM judgments aggregated via Bradley--Terry~\citep{bradley1952rank}. On the judge side, \citet{zheng2023judging} and \citet{liusie2024pairwise} establish that pairwise LLM judgments align better with human preferences than pointwise scoring, and Chatbot Arena~\citep{chiang2024chatbotarena} demonstrates that Bradley--Terry maximum-likelihood estimation converts noisy pairwise votes into stable rankings at scale. OpenDeepThink lifts this machinery from offline model ranking to in-the-loop search: the BT score defines elites, routes feedback, and selects the final submission.

\paragraph{Concurrent work.}
Several concurrent efforts explore parallel test-time reasoning: Population-Evolve~\citep{zhang2025populationevolve} maintains a population but aggregates by majority voting; ParaThinker~\citep{wen2025parathinker}, PaCoRe~\citep{hu2026pacore}, and Multiverse~\citep{yang2025multiverse} train models for parallel reasoning paths; SSA~\citep{qi2025ssa} trains a compact aggregator; V$_1$~\citep{singh2026v_1} shows that pairwise self-verification substantially outperforms pointwise scoring and proposes a tournament-based ranking for parallel candidate selection; Squeeze Evolve~\citep{maheswaran2026squeeze} addresses diversity collapse and cost efficiency in verifier-free evolution through multi-model orchestration; PDR~\citep{madaan2025rethinking} generates parallel drafts and distills them into a shared workspace for refinement; \citet{dang2026cognitivewell} design a competition-math pipeline with conjecture extraction to escape grader failure. OpenDeepThink is distinguished by being training-free, verifier-free, and applicable to open-ended outputs, with Bradley--Terry aggregation over pairwise critiques providing the selection signal.

\section{Method}
\label{sec:method}

\subsection{Setup}
\label{sec:overview}

\paragraph{Problem setup.} We study the problem of maximizing an LLM's accuracy on hard reasoning tasks under three simultaneous constraints: (i)~a fixed per-problem compute budget measured in API calls, (ii)~a wall-clock time limit, and (iii)~no access to a ground-truth verifier or hidden test cases at inference time. Only the problem statement is available; the solver has no access to web search, external knowledge retrieval, or tool use.
The wall-clock constraint is load-bearing: it rules out methods that merely extend a single reasoning chain, since such methods cannot convert additional compute into reduced response time.

\paragraph{Notation.} We denote a problem instance by $x$ and write $\mathcal{Y}^{(t)} = \{y^{(t)}_1, \dots, y^{(t)}_n\}$ for the population of $n$ candidate solutions at generation $t \in \{0, 1, \dots, T\}$. The same LLM $\pi$ serves as both generator and judge. For a pair $(y_i, y_j)$ at generation $t$, the judge produces an outcome $c^{(t)}_{ij} \in \{i \succ j,\; j \succ i,\; \text{tie}\}$ together with a pair of natural-language rationales $(r^{(t)}_{ij}, r^{(t)}_{ji})$, where $r^{(t)}_{ab}$ is the rationale supporting candidate $a$ over $b$. We aggregate the comparisons within generation $t$ into a Bradley--Terry score vector $\mathbf{s}^{(t)} \in \mathbb{R}^n$. We write $\mathcal{E}^{(t)} \subset \mathcal{Y}^{(t-1)}$ for the subset of candidates preserved as elites at generation $t$, and $\mathcal{D}^{(t)} \subset \mathcal{Y}^{(t-1)}$ for the subset discarded. We write $\hat y \in \mathcal{Y}^{(T)}$ for the solution selected after the final round. The pipeline has four hyperparameters: population size $n$, per-generation comparisons per candidate $K$, number of evolution generations $T$, and final-round comparisons per candidate $M$.

\begin{algorithm}[t]
\caption{OpenDeepThink.}
\label{alg:evoreason}
\begin{algorithmic}[1]
\Require Problem $x$; LLM $\pi$ (generator and judge); hyperparameters $n$, $K$, $T$, $M$.
\Ensure Selected solution $\hat y$.
\State $\mathcal{Y}^{(0)} \gets \{y^{(0)}_i \sim \pi(\cdot \mid x)\}_{i=1}^{n}$ \Comment{initial parallel sampling}
\For{$t = 1, \dots, T$}
    \State $\mathcal{P}^{(t)} \gets$ random pairing assigning each $y^{(t-1)}_i$ to $K$ peers, uniformly
    \State $\{(c^{(t)}_{ij}, r^{(t)}_{ij}, r^{(t)}_{ji})\}_{(i,j) \in \mathcal{P}^{(t)}} \gets$ parallel judgments by $\pi$ with randomized presentation order
    \State $\mathbf{s}^{(t)} \gets \arg\max_{\mathbf{s}} \mathcal{L}_{\text{BT}}(\mathbf{s}; \{c^{(t)}_{ij}\}) - \tfrac{1}{2}\lambda \lVert \mathbf{s} \rVert_2^2$ \Comment{L-BFGS}
    \State $\mathcal{E}^{(t)} \gets \text{top-}\lceil n/4 \rceil \text{ of } \mathcal{Y}^{(t-1)} \text{ by } \mathbf{s}^{(t)}$ \Comment{elite preservation}
    \State $\mathcal{D}^{(t)} \gets \text{bottom-}\lfloor n/4 \rfloor \text{ of } \mathcal{Y}^{(t-1)} \text{ by } \mathbf{s}^{(t)}$ \Comment{discard}
    \For{each $y^{(t-1)}_i \notin \mathcal{D}^{(t)}$ in parallel}
        \State $y^{(t)}_i \sim \pi(\cdot \mid x, y^{(t-1)}_i, \text{Aggregate}(\{r^{(t)}_{ji}\}_j))$ \Comment{feedback-driven mutation}
    \EndFor
    \State $\mathcal{Y}^{(t)} \gets \mathcal{E}^{(t)} \cup \{y^{(t)}_i\}_{i : y^{(t-1)}_i \notin \mathcal{D}^{(t)}}$
\EndFor
\State $\mathcal{P}^{\star} \gets$ random pairing assigning each $y \in \mathcal{Y}^{(T)}$ to $M$ peers, uniformly
\State $\{c^{\star}_{ij}\}_{(i,j) \in \mathcal{P}^{\star}} \gets$ parallel judgments by $\pi$
\State $\mathbf{s}^{\star} \gets \arg\max_{\mathbf{s}} \mathcal{L}_{\text{BT}}(\mathbf{s}; \{c^{\star}_{ij}\}) - \tfrac{1}{2}\lambda \lVert \mathbf{s} \rVert_2^2$
\State \Return $\hat y \gets \arg\max_{y \in \mathcal{Y}^{(T)}} s^{\star}_y$
\end{algorithmic}
\end{algorithm}

\subsection{Selection}\label{sec:pairwise}

\paragraph{Pairwise comparison design.}
At generation $t$, each candidate $y^{(t-1)}_i$ is compared against $K=4$ randomly sampled peers. The comparison prompt instructs the judge to identify the solution more likely to be accepted by a hypothetical online judge, to declare a tie when appropriate, and to supply a brief natural-language rationale $r^{(t)}_{ij}$ for each side; these rationales are reused as feedback during mutation. To mitigate position bias~\citep{zheng2023judging}, we randomize the presentation order in every comparison. We use the same model for generation and judgment: sharing the model reduces system complexity and demonstrates that the framework does not depend on an auxiliary verifier. Comparisons across different pairs are independent and execute in parallel.

\paragraph{Bradley--Terry aggregation.}
Given the pairwise outcomes $\{c^{(t)}_{ij}\}$ within a generation, we fit the Bradley--Terry (BT) score vector $\mathbf{s}^{(t)}$~\citep{bradley1952rank} under
\begin{equation}\label{eq:bt}
P(i \succ j) = \sigma(s^{(t)}_i - s^{(t)}_j),
\end{equation}
where $\sigma$ is the logistic sigmoid and ties contribute half a win to each side. We estimate $\mathbf{s}^{(t)}$ by maximizing the regularized log-likelihood of the observed comparisons with L-BFGS~\citep{liu1989lbfgs}, adding a small $\ell_2$ penalty for numerical stability and to fix the global shift, since the BT log-likelihood is invariant under additive shifts of $\mathbf{s}^{(t)}$. We prefer BT over raw win rate because each candidate faces only a random subset of opponents; BT internalizes opponent strength, yielding an adjusted ranking that raw win rate cannot. This adjustment is particularly important at $K=4$, where sampling noise from the pairing distribution is non-negligible. The same BT formulation underlies large-scale preference evaluation of LLMs~\citep{chiang2024chatbotarena}, and we exploit it here as a soft verifier that distills noisy pairwise signals into a coherent global ranking.

\paragraph{Implementation details.} We set $\lambda = 0.01$ with the penalty term $\tfrac{1}{2}\lambda\lVert\mathbf{s}\rVert_2^2$ for all BT fits. Pairwise comparisons sample a random $K$-regular matching without self-pairs; duplicate pairs within a generation are not permitted. Invalid JSON judge outputs are retried once; remaining failures are treated as ties.

\paragraph{Elite preservation.} After fitting the BT scores $\mathbf{s}^{(t)}$, the top $25\%$ of candidates are preserved as elites $\mathcal{E}^{(t)}$ and carried forward unchanged, while the bottom $25\%$ are discarded. The top $75\%$ (elites included) are routed to mutation, so each elite contributes both its original and a mutated variant. The bottom quartile is consistently worse than the middle range, so mutating it offers no benefit. Allowing elites to be mutated alongside the middle ranks maintains diversity without sacrificing selection pressure.

\subsection{Mutation}\label{sec:mutation}

The natural-language rationales $\{r^{(t)}_{ji}\}_j$ produced during pairwise comparison are a byproduct of selection that we deliberately recycle rather than discard. For each non-discarded candidate $y^{(t-1)}_i$ (the top $75\%$, including elites), we aggregate all feedback directed at it across its $K$ comparisons, yielding a targeted critique grounded in head-to-head failures. The mutator is conditioned on three inputs: (i)~the original problem statement, (ii)~the current solution, and (iii)~the aggregated feedback. The prompt permits the model to refine the existing solution or to abandon it for a fundamentally different approach. Informally, we observed that without this permission the mutator tends to apply local patches to a broken strategy rather than restructure it; we do not test this directly with a controlled ablation. Granting license to restart allows mutation to traverse qualitatively different solution sketches rather than hill-climbing within the neighborhood of a flawed initialization. Our mutation is executed as a synchronous parallel batch: all non-discarded candidates within a generation are mutated simultaneously and independently, each conditioned on its own aggregated feedback. Combined with elite preservation and bottom-quartile discarding, this yields a dynamic in which strong solutions are both retained and mutated, moderate ones are regenerated, and the weakest are eliminated.

\section{Experiments}
\label{sec:experiments}

\subsection{Setup}\label{sec:setup}

\paragraph{Benchmarks.} The primary evaluation suite pairs two complementary sources of competition-level programming problems totaling $192$ items, each requiring a correct C++ solution with no access to hidden test cases at inference time. CF-73 consists of $73$ recent Codeforces problems from Div.~1 and Div.~2 rounds rated roughly $2000$ to $3100$, annotated by an International Grandmaster and judged against the official Codeforces test suites~\citep{codeforces}; our local judge agrees with the official verdict on $99\%$ of submissions. All 73 problems are drawn from rounds held after August 2025, postdating the knowledge cutoff of every model evaluated (Gemini 2.5 Pro, 3 Flash, and 3.1 Pro), making pretraining contamination unlikely. The 1\% of disagreements between local and official verdicts are exclusively near-threshold TLE cases caused by differences in machine speed between our evaluation hardware and the Codeforces judging servers. NOI-119 consists of $119$ problems from a national informatics olympiad training program whose statements are public but whose tests and evaluation infrastructure remain private, accessed through a hidden online judge that returns binary accept/reject verdicts. Cross-domain experiments use $82$ questions sampled uniformly at random from the HLE gold set~\citep{phan2025hle}, a verified subset of the Humanity's Last Exam benchmark covering mathematics, natural sciences, computer science and AI, humanities, and social sciences.

\paragraph{Difficulty tiers.} We stratify the $192$ programming problems by the gen-$0$ pass@$1$ rate of the main-experiment base model, estimated from the $20$ independent samples that seed evolution. The cutoffs are the $33$rd and $67$th percentiles of that distribution and yield an Easy tier of $53$ problems with pass@$1 > 95\%$, a Medium tier of $75$ problems with $35\% < \text{pass@}1 \leq 95\%$, and a Hard tier of $64$ problems with pass@$1 \leq 35\%$. The source composition varies sharply with difficulty: Easy draws $32$ problems from CF and $21$ from NOI, Medium splits $31$/$44$, and Hard concentrates in NOI with only $10$ of its $64$ problems coming from CF. Within each tier, random pass@$1$ measures how often a single unranked sample succeeds, while BT top-$1$ measures whether the Bradley--Terry winner is accepted; the contrast between the two isolates aggregation's contribution from raw sampling coverage.

\paragraph{Models.} The main experiments use Gemini~3.1~Pro~\citep{comanici2025gemini} as both the generator and the judge. We additionally report Gemini~3~Flash and Gemini~2.5~Pro under identical hyperparameters and no per-model tuning. Pairwise judgments randomize presentation order throughout.

\paragraph{Compute budget.} The per-problem LLM-call budget decomposes into three contributions: the initial sampling of $n$ candidates, $T$ evolution generations each combining pairwise comparisons with mutation of the top $75\%$ (including elites), and a single denser final round of pairwise comparisons for selection. Summing these components gives
\begin{equation}
\label{eq:budget}
\text{Total}(n, K, T, M) \;=\; \underbrace{n\vphantom{\tfrac{n}{2}}}_{\text{sampling}} \;+\; T \cdot \Big( \underbrace{\tfrac{nK}{2}}_{\text{compare}} + \underbrace{\tfrac{3n}{4}}_{\text{mutate}} \Big) \;+\; \underbrace{\tfrac{nM}{2}}_{\text{final select}},
\end{equation}
which, at the main-experiment setting $n=20$, $K=4$, $T=3$, $M=10$, evaluates to approximately $285$ calls per problem. All four hyperparameters are held fixed across problems and benchmarks. The pipeline's sequential depth is eight LLM calls: one initial sampling round, two rounds (compare + mutate) per evolution generation for $T=3$ generations, and one final comparison round. All calls within each round are embarrassingly parallel across candidates. At the observed median per-call latency of approximately $200$ seconds for Gemini~3.1~Pro on competition-programming problems, this sequential depth corresponds to roughly $27$ minutes under full parallelization. Our implementation waits for all responses within each round; in deployment, over-provisioning parallel requests and discarding late completions could approach this bound without harming selection quality, since we observe that shorter-latency responses tend to have higher accept rates on our benchmarks. The sequential depth remains fixed at eight rounds regardless of population size $n$ or comparison count $K$.

\subsection{Main Result}\label{sec:main}

\Cref{tab:main-result} reports per-tier results on the full \dat{$192$}-problem benchmark.
The three difficulty tiers cleanly separate the two mechanisms of OpenDeepThink.
On Easy and Medium problems, where random pass@$1$ is already high, BT aggregation alone, applied to the unevolved gen-$0$ population, reaches saturation on Easy (\dat{$100\%$}) and near-saturation on Medium (\dat{$95\%$}); evolution adds almost nothing because there is little to fix.
The Hard tier is where evolution becomes load-bearing: random pass@$1$ climbs from \dat{$11\%$} to \dat{$36\%$} across three generations, and BT top-$1$ rises from \dat{$23\%$} to \dat{$50\%$} (95\% CI of gain: $[16, 39]$ pp, paired bootstrap).
The \dat{$+39$}-point gain over random pass@$1$ decomposes into \dat{$+25$} from evolution producing better candidates and \dat{$+14$} from selection picking better among them.
Aggregation dominates when the base model already solves most problems; evolution dominates at the capability frontier.
On CF-73, translating BT top-$1$ verdicts into effective Codeforces Elo (\Cref{app:elo}), a single gen-$0$ random pick yields \dat{$2851$} (95\% CI: $[2716, 2974]$) and the post-evolution BT winner attains \dat{$3256$} ($[3049, 3655]$), a \dat{$+405$}-Elo gain comparable to the \dat{$+411$} that Gemini~3~Deep Think achieves over Gemini~3.1~Pro on LiveCodeBench Pro~\citep{zheng2025livecodebenchpro}.
Because the two benchmarks differ in problem composition, we compare gains rather than absolute ratings.
\begin{table}[t]
  \centering
  \caption{OpenDeepThink on the CF-73 + NOI-119 benchmark with Gemini~3.1~Pro at $n{=}20$, $K{=}4$, $T{=}3$, $M{=}10$. Pass@$1$ is the empirical accept rate of a single unranked gen-$0$ sample and lower-bounds what naive sampling achieves. Oracle, shown in gray, is the gen-$0$ pass@$20$ score: it counts problems where at least one of the $20$ initial candidates is accepted and upper-bounds any gen-$0$ selector. Ours reports Bradley--Terry top-$1$ accuracy at each generation. Generations $0$--$2$ use the sparse $K{=}4$ comparisons from the evolution loop; Generation $3$ uses the final dense $M{=}10$ round. All comparisons fall within the per-problem budget of ${\sim}285$ calls; no additional comparisons are used for intermediate reporting.%
  }
  \label{tab:main-result}
  \vspace{6pt}
  \small
  \begin{tabular}{l cc @{\quad} cccc}
    \toprule
                                & NOI-$119$       & CF-$73$        & Easy           & Medium         & Hard           & All             \\
    \midrule
    \# Problems                 & \dat{$119$}     & \dat{$73$}     & \dat{$53$}     & \dat{$75$}     & \dat{$64$}     & \dat{$192$}     \\
    \addlinespace[2pt]
    Pass@$1$                    & \dat{$49\%$}    & \dat{$80\%$}   & \dat{$100\%$}  & \dat{$76\%$}   & \dat{$11\%$}   & \dat{$61\%$}    \\
    \textcolor{gray}{Oracle}    & \textcolor{gray}{$76\%$}  & \textcolor{gray}{$96\%$}  & \textcolor{gray}{$100\%$}  & \textcolor{gray}{$100\%$} & \textcolor{gray}{$50\%$}  & \textcolor{gray}{$83\%$}  \\
    \addlinespace[2pt]
    \textbf{OpenDeepThink} (Ours) &             &                &                &                &                &                 \\
    \quad Generation $0$        & \dat{$61\%$}    & \dat{$92\%$}   & \dat{$100\%$}  & \dat{$95\%$}   & \dat{$23\%$}   & \dat{$72\%$}    \\
    \quad Generation $1$        & \dat{$70\%$}    & \dat{$96\%$}   & \dat{$100\%$}  & \dat{$97\%$}   & \dat{$42\%$}   & \dat{$80\%$}    \\
    \quad Generation $2$        & \dat{$70\%$}    & \dat{$95\%$}   & \dat{$100\%$}  & \dat{$96\%$}   & \dat{$42\%$}   & \dat{$79\%$}    \\
    \quad Generation $3$        & \dat{$75\%$}    & \dat{$96\%$}   & \dat{$100\%$}  & \dat{$99\%$}   & \dat{$50\%$}   & \dat{$83\%$}    \\
    \bottomrule
  \end{tabular}
\end{table}

\paragraph{Cross-model transfer.}
\Cref{tab:cross-model} runs the same pipeline with identical hyperparameters on Gemini~3~Flash and Gemini~2.5~Pro, each evaluated on the difficulty tier matched to its sweet spot.
All three models improve on both pass@$1$ and BT top-$1$, and the balance between the two mechanisms shifts predictably along the capability axis: Gemini~3~Flash, the weakest model, shows its largest lift in pass@$1$ (\dat{$+30$}), meaning evolution is the dominant contributor; Gemini~3.1~Pro, the strongest, shows its largest lift in BT top-$1$ (\dat{$+27$}), meaning selection contributes more.
No per-model tuning is applied.
Two observations discipline the claim: first, evolution and selection contribute unequally but both contribute positively across all three models; second, problems that no model ever solves at gen-$0$ are rarely rescued. Of the \dat{seventeen} unsolved problems across the Flash and 2.5~Pro runs, none crosses \dat{$5\%$} pass@$1$ in any generation, consistent with the hypothesis that OpenDeepThink amplifies partial competence rather than inducing new capabilities from scratch; we do not test this directly.

\begin{table}[t]
  \centering
  \caption{Cross-model and cross-domain transfer. \textbf{(a)}~Cross-model generalization on CF-73 + NOI-119 difficulty tiers. Tier labels are defined relative to 3.1~Pro's gen-$0$ pass@$1$ (\Cref{sec:setup}). Hyperparameters $n{=}20$, $K{=}4$, $T{=}3$, $M{=}10$ are held fixed with no per-model tuning. \textbf{(b)}~HLE BT top-$1$ accuracy by category, gen-$0$ versus gen-$2$. Objective-correctness domains show directional gains; subjective-judgment domains show directional declines. Two singleton categories (engineering, chemistry) are omitted as underpowered. All models are from the Gemini family.%
  }
  \label{tab:cross-model}
  \vspace{6pt}
  \small
  \begin{minipage}[t]{0.46\textwidth}
    \centering
    \textbf{(a)} Cross-model\\[4pt]
    \footnotesize
    \begin{tabular}{l ccc}
      \toprule
                            & 3.1~Pro        & 3~Flash        & 2.5~Pro        \\
      \midrule
      Tier                  & Hard           & Medium         & Easy           \\
      \# Problems           & \dat{$64$}     & \dat{$75$}     & \dat{$53$}     \\
      gen-$0$ pass@$1$      & \dat{$11\%$}   & \dat{$39\%$}   & \dat{$28\%$}   \\
      gen-$0$ BT top-$1$    & \dat{$23\%$}   & \dat{$64\%$}   & \dat{$42\%$}   \\
      Final pass@$1$        & \dat{$36\%$}   & \dat{$69\%$}   & \dat{$50\%$}   \\
      Final BT top-$1$      & \dat{$50\%$}   & \dat{$76\%$}   & \dat{$60\%$}   \\
      \midrule
      $\Delta$ (pass / BT)  & \dat{$+25/27$} & \dat{$+30/12$} & \dat{$+22/18$} \\
      \bottomrule
    \end{tabular}
  \end{minipage}
  \hfill
  \begin{minipage}[t]{0.50\textwidth}
    \centering
    \textbf{(b)} Cross-domain (HLE)\\[4pt]
    \footnotesize
    \begin{tabular}{lcccc}
      \toprule
      Category & $N$ & gen-$0$ & gen-$2$ & $\Delta$ \\
      \midrule
      Mathematics                   & \dat{$34$} & \dat{$56\%$} & \dat{$61\%$} & \dat{$+5$}  \\
      Biology / Medicine            & \dat{$7$}  & \dat{$43\%$} & \dat{$57\%$} & \dat{$+14$} \\
      Physics                       & \dat{$6$}  & \dat{$33\%$} & \dat{$50\%$} & \dat{$+17$} \\
      Computer Science / AI         & \dat{$15$} & \dat{$67\%$} & \dat{$64\%$} & \dat{$-2$}  \\
      Humanities / Soc.\ Sci.       & \dat{$8$}  & \dat{$75\%$} & \dat{$50\%$} & \dat{$-25$} \\
      Other                         & \dat{$10$} & \dat{$70\%$} & \dat{$40\%$} & \dat{$-30$} \\
      \midrule
      All                           & \dat{$82$} & \dat{$58.5\%$} & \dat{$54.3\%$} & \dat{$-4$} \\
      \bottomrule
    \end{tabular}
    \label{tab:hle-category}
  \end{minipage}
\end{table}

\paragraph{Cross-domain transfer.}\label{sec:hle}
We evaluate on \dat{$82$} HLE questions with $n{=}12$, $K{=}4$, $T{=}2$, $M{=}10$.
The key contrast with competitive programming is immediate: BT top-$1$ accuracy declines from \dat{$58.5\%$} to \dat{$54.3\%$} across two generations, while majority vote rises by \dat{$+3.1$} points (\Cref{fig:hle-overall}).
The category breakdown (\Cref{tab:cross-model}b) suggests a directional pattern, though category-level sample sizes ($N{=}6$--$8$ outside mathematics) preclude strong statistical claims.
Domains with objectively verifiable answers, namely mathematics, biology, and physics, show directional BT top-$1$ gains of \dat{$+5$} to \dat{$+17$} points, directionally consistent with competitive programming.
Domains with ambiguous correctness criteria, humanities and social sciences, decline by \dat{$25$} to \dat{$30$} points.
Whether OpenDeepThink helps or hurts tracks the reliability of pairwise LLM judgment: where the judge can discriminate correct from incorrect, BT selection amplifies signal; where it cannot, iterative selection amplifies noise.

\section{Discussion}\label{sec:discussion}

\begin{figure}[t]
  \centering
  \begin{subfigure}{0.48\columnwidth}
    \centering
    \includegraphics[width=\linewidth]{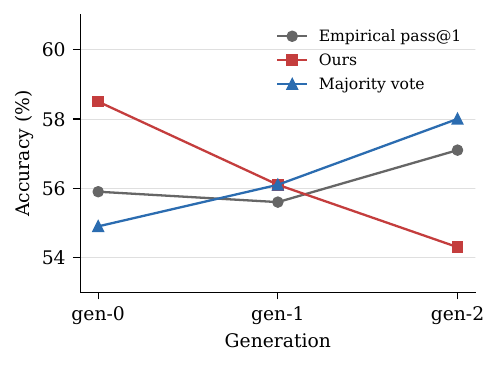}
    \caption{HLE aggregate, $N{=}\dat{82}$.}
    \label{fig:hle-overall}
  \end{subfigure}
  \hfill
  \begin{subfigure}{0.48\columnwidth}
    \centering
    \includegraphics[width=\linewidth]{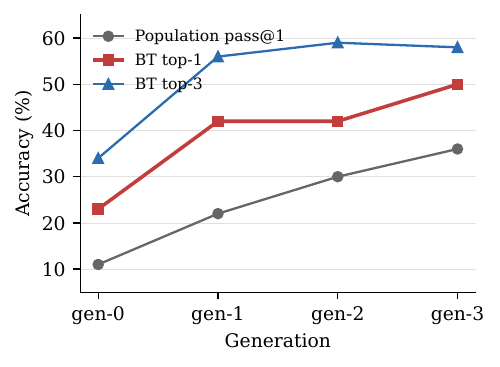}
    \caption{Iteration ablation on Hard, $N{=}\dat{64}$.}
    \label{fig:iter-ablation}
  \end{subfigure}
  \caption{\textbf{(a)}~HLE aggregate accuracy across evolution generations. Three aggregators are tracked: pass@$1$, BT top-$1$ (Ours), and majority vote. BT top-$1$ is the only aggregator that declines, consistent with the pairwise judge's unreliability on subjective domains (\Cref{tab:cross-model}b). \textbf{(b)}~Iteration dynamics on the Hard tier ($N{=}\dat{64}$, pairwise $K{=}4$). Population pass@$1$ and BT top-$k$ are shown per generation; gen-$3$ values use the final dense BT round.}
  \label{fig:line-charts}
\end{figure}

\paragraph{Pairwise selection outperforms pointwise under matched generation quality.}
The advantage of OpenDeepThink's Bradley--Terry selector over pointwise scoring is not an artifact of better candidates; it persists after generation quality is equalized and traces entirely to the selection mechanism.
The root cause is positive bias in pointwise scoring: pointwise judgment achieves high recall on accepted solutions but poor recall on wrong ones, so it cannot reliably reject incorrect candidates~\citep{zheng2023judging,liusie2024pairwise}.
Pairwise comparison sidesteps this failure mode by reducing judgment to a relative contrast that requires no calibrated quality threshold.
On a \dat{$500$}-pair diagnostic drawn from NOI-119 (\Cref{app:pairwise-diagnostic}), pairwise judgment reaches \dat{$86\%$} accuracy versus \dat{$59\%$} for pointwise.
This finding aligns with \citet{singh2026v_1}, who independently demonstrate that pairwise self-verification is a substantially more accurate primitive than pointwise scoring for selecting among parallel candidates.
The gap survives two increasingly aggressive controls for generation quality.
First, applying \dat{$14$} pointwise judgments per candidate to the same unevolved gen-$0$ pool yields only \dat{$26\%$} top-$1$ on the Hard tier at a budget of \dat{$300$} calls (20 shared initial samples $+$ 280 pointwise judgments).
Second, giving the baseline its own refinement budget, six rounds of Self-Refine~\citep{madaan2023selfrefine} followed by \dat{$8$} pointwise votes per candidate (\dat{$300$} calls), closes the generation gap almost entirely (pass@$1$: \dat{$33\%$} vs.\ OpenDeepThink's \dat{$36\%$}), yet top-$1$ reaches only \dat{$41\%$} against \dat{$50\%$}: the \dat{$9$}-point residual is pure selection signal.
The effect is not merely average-case. Of the \dat{$64$} Hard problems, OpenDeepThink's BT top-$1$ is accepted on \dat{$32$}; the strongest pointwise baseline is accepted with certainty---all tied-top candidates correct---on only \dat{$6$}. OpenDeepThink uniquely solves \dat{$27$} problems that the baseline never reliably solves, while the baseline uniquely solves \dat{$1$}. The baseline's \dat{$41\%$} figure reflects the expected accept rate under random tie-breaking among equally-scored candidates, not a deterministic solve count.

\paragraph{Negative feedback carries nearly all the mutation signal.}
Within a single mutation round, the improvement is almost entirely driven by negative feedback; positive feedback is statistically indistinguishable from no feedback at all (\Cref{tab:feedback-ablation}).
Telling the mutator what went wrong carries actionable signal; telling it what went right adds nothing beyond what the model already infers from seeing its own solution.
Structuring the negative signal as pairwise critique at $K{=}4$ nearly doubles the net rescue rate over the no-feedback baseline, because head-to-head contrast surfaces failure modes that a single trajectory cannot self-diagnose, consistent with the discriminative advantage that motivates pairwise selection in the first place.
Beyond $K{=}4$ the return reverses: $K{=}5$ regresses as the mutator receives more contrasts than it can integrate in a single rewrite\footnote{The primary role of $K$ is to supply enough pairwise observations for reliable Bradley--Terry ranking over $n{=}20$ candidates; the feedback signal recycled for mutation is a byproduct of this comparison budget. Because the ranking-precision constraint is statistical rather than model-dependent, we expect the operating point to be relatively stable across judge models, though we have not explicitly verified this.}.
Stratifying by difficulty sharpens the picture: on Medium problems, pairwise feedback's rescue advantage over no feedback is \dat{$\sim\!26$} points; on Hard problems it narrows to \dat{$\sim\!4$} points.
Feedback helps the mutator cross a nearby acceptance threshold, not learn a fundamentally new algorithm. Evolution amplifies partial competence.

\clearpage
\begin{table}[t]
  \centering
  \caption{Feedback strategy ablation on \dat{$500$} solutions from \dat{$64$} NOI problems. AC and WA denote the online judge's Accepted and Wrong Answer verdicts; \dat{$194$} solutions are originally AC and \dat{$306$} originally WA. One round of mutation is applied per strategy. Rescue counts WA $\to$ AC transitions; Degradation counts AC $\to$ WA; $\Delta$ is rescued minus degraded. The chosen setting is \textbf{pairwise $K{=}4$}.}
  \label{tab:feedback-ablation}
  \begin{tabular}{lcccc}
    \toprule
    Feedback strategy & AC rate & Rescued (WA$\to$AC) & Degraded (AC$\to$WA) & $\Delta$ \\
    \midrule
    No feedback            & \dat{$47.6\%$} & \dat{$48$ \;\;($15.7\%$)} & \dat{\phantom{0}$4$ \;\;($2.1\%$)} & \dat{$+44$} \\
    Positive only          & \dat{$47.8\%$} & \dat{$52$ \;\;($17.0\%$)} & \dat{\phantom{0}$7$ \;\;($3.6\%$)} & \dat{$+45$} \\
    Negative only          & \dat{$49.4\%$} & \dat{$68$ \;\;($22.2\%$)} & \dat{$15$ \;\;($7.7\%$)}           & \dat{$+53$} \\
    Pairwise, all pairs    & \dat{$50.4\%$} & \dat{$70$ \;\;($22.9\%$)} & \dat{$12$ \;\;($6.2\%$)}           & \dat{$+58$} \\
    Pairwise, $K{=}2$      & \dat{$51.6\%$} & \dat{$73$ \;\;($23.9\%$)} & \dat{\phantom{0}$9$ \;\;($4.6\%$)} & \dat{$+64$} \\
    Pairwise, $K{=}3$      & \dat{$51.2\%$} & \dat{$69$ \;\;($22.5\%$)} & \dat{\phantom{0}$7$ \;\;($3.6\%$)} & \dat{$+62$} \\
    \textbf{Pairwise, $K{=}4$} & \dat{\textbf{$52.4\%$}} & \dat{\textbf{$74$ \;\;($24.2\%$)}} & \dat{\textbf{\phantom{0}$6$ \;\;($3.1\%$)}} & \dat{$\mathbf{+68}$} \\
    Pairwise, $K{=}5$      & \dat{$50.8\%$} & \dat{$68$ \;\;($22.2\%$)} & \dat{\phantom{0}$8$ \;\;($4.1\%$)} & \dat{$+60$} \\
    \bottomrule
  \end{tabular}
\end{table}

\begin{wrapfigure}{r}{0.5\columnwidth}
  \vspace{-1.2em}
  \centering
  \includegraphics[width=\linewidth]{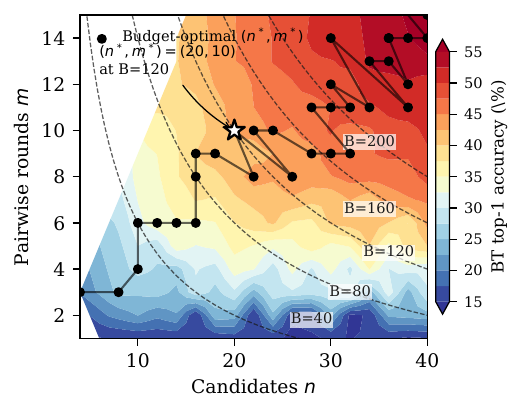}
  \caption{BT-only scaling on \dat{$10$} Hard problems. Color is BT top-$1$ accuracy on the $(n, m)$ plane from Monte-Carlo simulation (\dat{$500$} trials per cell, \dat{$40$} pre-judged candidates per problem). Dashed contours mark equal-budget allocations. Black dots trace the budget-optimal $(n^*, m^*)$; the white star marks $(20, 10)$ at $B{=}120$, matching the main pipeline's selection-only budget (excluding evolution calls). Experimental details in \Cref{app:bt-scaling-setup}.}
  \label{fig:bt-scaling}
  \vspace{-1em}
\end{wrapfigure}
\paragraph{Evolution is front-loaded; dense comparison extracts the residual.}
Across generations, the largest single improvement occurs at gen-$0 \to$ gen-$1$, where the first mutation round converts the most tractable failures (\Cref{fig:iter-ablation}). Meanwhile, sparse intra-generation comparison ($K{=}4$) saturates by gen-$2$: once candidates become roughly comparable, a low comparison budget can no longer resolve their ranking.
The final dense round ($M{=}10$) breaks this plateau, extracting an additional \dat{$+8$} points at top-$1$ beyond sparse gen-$2$ BT.
Evolution and dense selection therefore play distinct and complementary roles: evolution raises the ceiling of the candidate pool; dense Bradley--Terry extracts it.
The quadratic cost of comparison ($m^* \approx n/2$ yields $\Theta(n^2)$ total) makes population size the binding constraint; \Cref{fig:bt-scaling} confirms that the budget-optimal allocation at $B{=}\dat{120}$ matches the main pipeline's $(n, M) = (\dat{20}, \dat{10})$.

\section{Conclusion}
\label{sec:conclusion}

OpenDeepThink addresses the selection bottleneck that emerges when test-time compute scaling is parallelized rather than deepened. Instead of extending a single chain of thought, the framework maintains a population of $n$ candidates and evolves them over $T$ generations. The same LLM acts as both generator and pairwise judge: Bradley--Terry aggregation yields a soft verifier, and the comparison critiques drive feedback-directed mutation. The pipeline runs at a sequential depth of eight LLM calls. On Codeforces, this lifts Gemini~3.1~Pro's effective Elo by \dat{$+405$} points, and the same hyperparameters transfer to Gemini~3~Flash and Gemini~2.5~Pro without retuning. On HLE, gains hold where pairwise judgment is reliable and reverse where it is not, so the soft verifier is only as good as the comparisons it aggregates.

\paragraph{Limitations.}
The framework has been validated only on Gemini-family models; whether the gains transfer to architecturally different LLMs is unknown.
The per-problem cost of ${\sim}285$ API calls is substantial and may be prohibitive for latency-sensitive applications.
The soft verifier inherits the judge's biases: on HLE domains where pairwise judgment is unreliable, evolution actively degrades accuracy.
Finally, the $25\%$ elite ratio and the license-to-abandon prompt were chosen by informal tuning rather than controlled ablation.


\clearpage
\bibliography{references}

\clearpage
\beginsupplement

\section{Technical appendices and supplementary material}

\subsection{HLE category breakdown}
\label{app:hle-category}

The HLE category breakdown is reported in \Cref{tab:cross-model}(b) in the main text. Two singleton categories, engineering and chemistry, are omitted as underpowered.

\subsection{Elo Rating Estimation}\label{app:elo}

The effective Elo ratings of \Cref{sec:main} are computed on the CF-$73$ subset, since NOI problems do not have published Codeforces ratings. We adopt the standard Elo logistic model in which the probability that a player at rating $R_{\text{model}}$ solves a problem at rating $R_{\text{problem}}$ is
\[
  P(\text{solve}) \;=\; \frac{1}{1 + 10^{(R_{\text{problem}} - R_{\text{model}})/400}}.
\]
Each problem's $R_{\text{problem}}$ is taken from the Codeforces website. We estimate $R_{\text{model}}$ by maximum a posteriori (MAP) under a Gaussian prior $\mathcal{N}(3100,\,500^2)$ centered loosely on the published rating of Gemini~3.1~Pro, optimizing the posterior with \texttt{scipy.optimize.minimize\_scalar} over the bounded interval $[1000,\,5000]$.

We report two scenarios. For gen-$0$ pass@$1$, the per-problem likelihood is Binomial with $n = 20$ independent gen-$0$ samples and $k$ accepted; this measures the rating implied by naive sampling. For the final BT top-$1$, the per-problem likelihood is Bernoulli, treating the BT-ranked top candidate as a single submission that is either accepted or rejected; this measures the rating of the post-evolution selector. The same prior, optimizer, and per-problem ratings are used in both scenarios.

Confidence intervals are obtained by bootstrap resampling. We draw $1000$ resamples by resampling problems with replacement (a problem contributes its full likelihood factor each time it is drawn), refit $R_{\text{model}}$ on each resample, and report the $2.5$ and $97.5$ percentiles as the $95\%$ CI. Resampling at the problem level captures the dominant source of variance, since per-problem outcomes are the noisy units; within-problem sample variance is already absorbed into the Binomial likelihood for the gen-$0$ scenario (this absorption does not apply to the BT top-$1$ scenario, which is per-problem Bernoulli).

\subsection{Pairwise diagnostic setup}\label{app:pairwise-diagnostic}

The \dat{$500$} (AC, WA) solution pairs used in the pairwise-vs-pointwise diagnostic of \Cref{sec:discussion} are drawn from \dat{$\sim\!60$} NOI-119 problems. Each problem contributes \dat{$20$} candidate solutions whose AC/WA labels are determined by the online judge against its private test suite, independently of any Bradley--Terry output. We form within-problem pairs from the $\binom{20}{2}$ combinations, retain only those containing one AC and one WA solution, and sample \dat{$500$} pairs uniformly across the resulting pool. The pointwise variant scores each solution in isolation; the pairwise variant compares the two. Same judge model, randomized presentation order~\citep{zheng2023judging}. Pointwise attains \dat{$96.4\%$} accuracy on AC but only \dat{$62.2\%$} on WA, yielding a joint correctness of \dat{$59.2\%$}. Pairwise reaches \dat{$86.2\%$}. Since AC/WA labels are determined by the online judge independently of any BT output, the diagnostic does not feed circularly into the main result.

\subsection{BT-only scaling study setup}\label{app:bt-scaling-setup}

The scaling study of \Cref{fig:bt-scaling} isolates Bradley--Terry aggregation from evolution on \dat{$10$} Hard problems with gen-$0$ AC rate $\in [1/20,\,4/20]$. For each problem we draw \dat{$40$} candidates from $\pi$, label every candidate against the public test suite, and pre-compute all \dat{$780$} pairwise judgments via full round-robin: \dat{$39$} rounds of \dat{$20$} pairs each. For each budget $B = n + m \cdot n/2$ and population size $n$, we Monte-Carlo sample $n$ candidates without replacement, simulate $m = \lfloor 2(B-n)/n\rfloor$ rounds of random-pairing comparison with \dat{$500$} trials per cell, aggregate with Bradley--Terry, and record whether the top-$1$ pick is accepted. Optimal $n$ grows sublinearly with $B$ at roughly $n^* \propto B^{0.6}$, so additional budget is better spent on broader sampling than on deeper comparison at the margin.

\subsection{Baseline Details}\label{app:baseline-details}

Section~\ref{sec:discussion} compares OpenDeepThink against two pointwise baselines. We document their implementation here.

\paragraph{Pointwise judge.} Each candidate solution is scored in isolation using the same base model (Gemini~3.1~Pro) at temperature $1.0$. The prompt asks whether the solution is correct for all valid inputs and requires a final line of exactly \texttt{VERDICT: YES} or \texttt{VERDICT: NO}. Responses that do not match this format are discarded. For each candidate, the judge is called $N$ times independently ($N{=}14$ for the standalone pointwise baseline, $N{=}8$ for the Self-Refine variant), and the candidate's score is the count of \texttt{YES} verdicts. The top-$1$ pick is drawn uniformly at random from the candidates with the highest \texttt{YES} count; the reported accuracy is the fraction of AC solutions in this tied-top set, equivalent to the expected accept rate under random tie-breaking.

\paragraph{Self-Refine baseline.} Starting from the same $20$ gen-$0$ candidates as OpenDeepThink, each candidate is independently refined for six rounds. Each round uses a single-step prompt that asks the model to review its current solution and either output it unchanged or produce an improved version; unlike the two-step feedback-then-refine protocol of \citet{madaan2023selfrefine}, our implementation merges feedback and rewriting into a single call. No cross-solution information is provided: each trajectory sees only the problem statement and its own most recent code. After six rounds, the $20$ refined candidates are scored with $8$ pointwise votes each and selected by the same \texttt{YES}-count procedure described above. The total budget is $300$ calls per problem ($20$ shared initial samples $+$ $120$ refinement $+$ $160$ pointwise), comparable to OpenDeepThink's $285$.

\subsection{Prompt Templates}\label{app:prompts}

The full pipeline code is available at \url{https://github.com/ZhouShang0817/open-deep-think}.

Three prompt templates govern the pipeline. Generation and mutation share the same system prompt. The judge uses no system prompt; its full instruction is in the user message. Temperature is $1.0$ for generation and mutation, $0.0$ for judging. All placeholders are literal substitutions with no additional formatting.

\paragraph{Generation.}
Used in gen-$0$ sampling to produce $n$ initial candidates per problem.

\begin{tcolorbox}[colback=gray!5, colframe=gray!60, fontupper=\small\ttfamily, title={\small\sffamily\bfseries System}, breakable]
You are an expert competitive programmer.\\
Output your solution as a single \textasciigrave\textasciigrave\textasciigrave cpp ... \textasciigrave\textasciigrave\textasciigrave{} block, preceded by brief reasoning.
\end{tcolorbox}

\begin{tcolorbox}[colback=gray!5, colframe=gray!60, fontupper=\small\ttfamily, title={\small\sffamily\bfseries User}, breakable]
\{problem\}
\end{tcolorbox}

\paragraph{Pairwise comparison.}
Used in per-generation comparison and the final dense BT round. Returns a JSON object with a winner verdict and per-side feedback. Presentation order of Solutions A and B is randomized per comparison to mitigate position bias.

\begin{tcolorbox}[colback=gray!5, colframe=gray!60, fontupper=\small\ttfamily, title={\small\sffamily\bfseries User (no system prompt)}, breakable]
You are a competitive programming expert.\\[4pt]
\#\# Problem Statement\\
\{problem\}\\[4pt]
\#\# Solution A\\
\textasciigrave\textasciigrave\textasciigrave cpp\\
\{code\_a\}\\
\textasciigrave\textasciigrave\textasciigrave\\[4pt]
\#\# Solution B\\
\textasciigrave\textasciigrave\textasciigrave cpp\\
\{code\_b\}\\
\textasciigrave\textasciigrave\textasciigrave\\[4pt]
Which solution is more likely to receive an Accepted verdict from an online judge --- meaning it produces correct output within the time and memory limits for all valid inputs?\\[4pt]
If both solutions appear incorrect (wrong answer, TLE, or other issues), choose the one that requires fewer modifications to become Accepted.\\[4pt]
If they are fundamentally identical or equally likely to be Accepted, output TIE.\\[4pt]
Respond with a JSON object and nothing else, in exactly this format:\\
\{\\
\quad"feedback\_a": "one sentence on Solution A's key strength or critical flaw",\\
\quad"feedback\_b": "one sentence on Solution B's key strength or critical flaw",\\
\quad"winner": "A or B or TIE"\\
\}
\end{tcolorbox}

\paragraph{Mutation.}
Used to mutate non-discarded candidates (top 75\%, including elites) each generation. The variant with feedback is the default; the variant without feedback is a fallback for candidates with no comparisons in the current generation.

\begin{tcolorbox}[colback=gray!5, colframe=gray!60, fontupper=\small\ttfamily, title={\small\sffamily\bfseries System}, breakable]
Same as generation.
\end{tcolorbox}

\begin{tcolorbox}[colback=gray!5, colframe=gray!60, fontupper=\small\ttfamily, title={\small\sffamily\bfseries User (with feedback)}, breakable]
\#\# Problem\\
\{problem\}\\[4pt]
\#\# Solution\\
\textasciigrave\textasciigrave\textasciigrave cpp\\
\{code\}\\
\textasciigrave\textasciigrave\textasciigrave\\[4pt]
\#\# Pairwise Feedback\\
This solution was compared against other solutions multiple times:\\[4pt]
\{feedback\_sections\}\\[4pt]
\#\# Task\\
Write a solution that maximizes the probability of Accepted.
You may refine the existing solution or take a different approach if the current one is fundamentally flawed.\\[4pt]
Think briefly, then output your final solution as a single \textasciigrave\textasciigrave\textasciigrave cpp ... \textasciigrave\textasciigrave\textasciigrave{} block.
\end{tcolorbox}

\noindent The \texttt{\{feedback\_sections\}} placeholder is filled with the candidate's pairwise critiques from the current generation, partitioned by outcome:

\begin{tcolorbox}[colback=gray!5, colframe=gray!60, fontupper=\small\ttfamily, title={\small\sffamily\bfseries Feedback section format}, breakable]
\#\#\# Wins (this solution was judged better):\\
- \{feedback\}\\[2pt]
\#\#\# Ties (judged equally likely to be Accepted):\\
- \{feedback\}\\[2pt]
\#\#\# Losses (this solution was judged worse):\\
- \{feedback\}
\end{tcolorbox}

\noindent Empty sections are omitted. Within each feedback string, the judge's references to ``Solution~A'' / ``Solution~B'' are rewritten to ``this solution'' / ``the other solution'' so the mutator receives self-relative critiques. The without-feedback variant omits the Pairwise Feedback section entirely; all other text is identical.

\end{document}